\documentclass[conference, compsoc]{IEEEtran}
\IEEEoverridecommandlockouts
\usepackage{amsmath,amssymb}
\usepackage{iftex}
\usepackage{stfloats}
\usepackage{float}
\usepackage{tabularx}
\usepackage{booktabs}
\usepackage[T1]{fontenc}
\usepackage[utf8]{inputenc}
\usepackage{textcomp} 

\usepackage{xcolor}
\usepackage{longtable,booktabs,array}
\usepackage{calc} 
\usepackage{etoolbox}

\IfFileExists{footnotehyper.sty}{\usepackage{footnotehyper}}{\usepackage{footnote}}
\makesavenoteenv{longtable}
\usepackage{graphicx}

\def\maxwidth{\ifdim\Gin@nat@width>\linewidth\linewidth\else\Gin@nat@width\fi}
\def\maxheight{\ifdim\Gin@nat@height>\textheight\textheight\else\Gin@nat@height\fi}
\makeatother
\setkeys{Gin}{width=\maxwidth,height=\maxheight,keepaspectratio}
\makeatletter
\def\fps@figure{htbp}
\makeatother
\ifLuaTeX
  \usepackage{selnolig}  
\fi
\IfFileExists{bookmark.sty}{\usepackage{bookmark}}{\usepackage{hyperref}}
\IfFileExists{xurl.sty}{\usepackage{xurl}}{} 
\hypersetup{
  hidelinks}

\title{\LARGE Object-based Probabilistic Similarity Evidence of Sparse Latent Features from Fully Convolutional Networks}
\author{Cyril Juliani$^{*}$
\thanks{$^{*}$Corresponding email: \href{mailto:cyril.juliani@gmail.com}{cyril.juliani@gmail.com}}%
}

\begin{document}
\maketitle

\begin{abstract}
Similarity analysis using neural networks has emerged as a powerful technique for understanding and categorizing complex patterns in various domains. By leveraging the latent representations learned by neural networks, data objects such as images can be compared effectively. This research explores the utilization of latent information generated by fully convolutional networks (FCNs) in similarity analysis, notably to estimate the visual resemblance of objects segmented in 2D pictures. To do this, the analytical scheme comprises two steps: (1) extracting and transforming feature patterns per 2D object from a trained FCN, and (2) identifying the most similar patterns through fuzzy inference. The step (2) can be further enhanced by incorporating a weighting scheme that considers the significance of latent variables in the analysis. The results provide valuable insights into the benefits and challenges of employing neural network-based similarity analysis for discerning data patterns effectively.
\end{abstract}

\section{1. Introduction}

Computer vision and image analysis have gained significant importance in
various research fields. Imagery data often contain multiple 2D objects
that need to be accurately depicted and organized, enabling the study of
unknown patterns or targeted object retrieval (e.g., {[}1, 2, 3{]}).
Similarity analysis plays a crucial role in such study. By comparing the
resemblance of image components, the identification of recurring
structures or visual relationships becomes possible. In practice, this
comparison can involve transforming the image components into a vector
space and utilizing specific metrics like distance, inner product, or
entropy measures {[}4{]} for comparing vector elements. While feature
vectors translate object properties from image sets, they can be
estimated via computational methods that learn to discern visual
features at the pixel level. Deep learning using neural networks has
particularly advanced this area, enabling remarkable progress in object
recognition across diverse aspects and distortions {[}5, 6{]}. The
generalization ability of such networks makes it possible to extract
spatial details into semantic information that can be exploited for
similarity analysis. Notably, Siamese networks have been widely used for
this task {[}7, 8, 9{]}. They learn similarity or dissimilarity between
pairs of inputs by using twin neural networks with shared weights to
extract features from two input samples, and then compare the feature
representations to compute a similarity or dissimilarity score. Triplet
network is another approach {[}10, 11{]} where embeddings are learnt
using triplets of samples consisting of an anchor, a positive (similar)
example, and a negative (dissimilar) example to train the network. While
training, the distance between similar samples is minimized and
dissimilar samples are maximized. Another method is deep metric
learning, which tackles similarity analysis tasks by learning a feature
embedding space where similar samples are closer to each other, and
dissimilar samples are farther apart {[}12, 13{]}. It typically involves
training a neural network to optimize a similarity or distance metric
loss function.

While these learning techniques utilize neural networks for similarity
analysis, they need to be optimized for distinct aspects of similarity.
In contrast, this research aims to demonstrate the effectiveness of
utilizing features extracted from pre-trained networks for similarity
analysis without relying on a specific loss criterion for such task. The
proposed analytical approach consists of two steps: (1) optimizing the
network\textquotesingle s feature representativeness during training,
and (2) leveraging the learned embeddings post-training via a
mathematical approach. Using a collection of images, the learning model
captures the visual properties of 2D objects identified in each image.
Techniques like increasing network sparsity at the channel level, i.e.,
reducing the number of network features, can be employed to enhance
feature quality during the model training. Subsequently, these features
are extracted, transformed, and compared through a mathematical
procedure.

One commonly used learning model is the fully convolutional network
(FCN) {[}14{]}, which can capture high-level abstractions of images via
convolutions. Since a convolution operation is a mathematic
approximation of the activity of a neuron (stimulus) within its
receptive field {[}6{]}, stimuli for each hidden unit can presumably
respond to an object(s) of interest in the original data space (noted
\(\mathcal{X}\)) that we eventually attempt to segment from the last
layer of FCNs. Stimuli patterns for a single prediction class can be
multiple, i.e., the network determines diverse feature patterns learned
for the corresponding class. An example is shown by {[}15{]} who
clustered various latent patterns from an FCN considering a one-class
segmentation problem i.e., a main object class, and a sub-class being a
chunk of the main one for detailing the similarity analysis and
resulting clusters. They demonstrated that a significant portion of the
encoding-decoding activation space of the trained FCN (noted
\(\mathcal{A}\)) exhibits multimodality. By leveraging the multimodality
of convolutional maps, new possibilities for object-oriented similarity
analysis can be unlocked.

This paper focuses on conducting a similarity analysis of objects
determined by FCNs. The approach complements the work presented in
{[}15{]}, which demonstrated the potential of leveraging feature
patterns learned from a supervised segmentation model for object
categorization. This can be done without learning a latent space
specifically for a similarity analysis (or clustering) task when mapping
\(\mathcal{X}\) to a new space. In this context, an FCN is trained to address a
one-class segmentation problem, and then, similarity of segmentation is
measured in a second step, conducted post-training. Results of this
segmentation are used to extract activated regions of network feature
maps associated to object segments. As described in {[}15{]} the
extraction procedure considers the backward propagation of a one-class
segmentation mask throughout the network to extract sub-regions of \(N\)
activated feature maps \(F = \left\{ f_{1},\ldots,f_{N} \right\}\)
where, layer-per-layer, representation sets
\(M = \left\{ m_{1},\ldots,m_{N} \right\}\), locally responsive to the
respective one-class segment in \(F\) are collected. On this basis,
{[}15{]} revealed groups of similar feature patterns by (1) taking the
mean value (or magnitude) of elements of \(M_{i}\) to compose a (mean)
activation vector \(v_{i} = (x_{i,1},\ldots,x_{i,N})\) associated to an
\(i\)th segmented object, then (2) comparing feature vectors of multiple
objects by correlation analyses, and (3) clustering hierarchically these
vectors based on correlation results. Resulting clusters, i.e., groups
of similar segments, or objects extracted from \(\mathcal{X}\), were evaluated
visually, but the measure of similarity from \(v_{i}\) among objects of
a same cluster was done in a Euclidean space. In this case, the
evaluation of similarity (in the semantic sense) shall require
estimating the distance of pairs of vectors for which we know the
meaning. This is a difficult task because a cluster is specified by
analogous latent features whose meanings are unknown. From a statistical
standpoint, it is however possible to quantify this similarity for more
interpretable and guided object identification. To address this, a
probabilistic similarity measure of \(v_{i}\) is proposed in this study,
by employing fuzzy partitioning of the latent domain \(\mathcal{A}\)
using fuzzy sets. Technically, the pattern of a reference object \(q\)
(query), denoted as \(v_{q}\), is collected, and the proximity of its
components to those of \(v_{i}\), for any object indexed \(i\) in a set
of segmented objects \(O\), is evaluated. An element
\(x_{i,j} \in v_{i}\) is thus evaluated in local neighborhoods of an
element \(x_{q,j} \in v_{q}\). As this neighborhood is fuzzy, we
determine the probabilistic ``distance'' between these elements using a
membership function (e.g., Gaussian function). The degree of fuzziness
of any feature vector \(v_{i}\), with respect to \(v_{q}\), forms the
basis of similarity search. An efficient way to explore analogous
patterns is to rank \(v_{i}\) by order of similarity
(\(\forall i \in O\)) with respect to a queried object indexed \(q\).

In this research, the objective is to tackle the calculation of
similarity scores between feature patterns through fuzzy inference. The
methodology is exemplified by utilizing a standard FCN, such as U-Net
{[}16{]}, which is capable of learning sparse semantic representations.
This model is trained using the UT Zappos50k datasets {[}17{]},
consisting of 12,833 images of boots, all of the same dimension and
orientation (see examples in Fig.1). Sparsity is generated at the
channel level, and implemented for dimensionality reduction of patterns
extracted from \(\mathcal{A}\), which are used in similarity
computations. A term enforcing sparsity is added to the loss to set
several (activated) convolution outputs to zero. After training, the
extraction and transformation procedures mentioned earlier are employed
to construct \(O\) and its corresponding feature vectors. Subsequently,
the "inactive" features, along with unimportant ones, are pruned using
eigen decomposition. The remaining "active" features (not pruned) are
processed for similarity computation of the feature vectors. The
approach will be optimized through the presentation of learning metrics.
Additionally, the similarity search scheme will be tested using features
extracted from a non-FCN model, ResNet50 {[}18{]}, which is trained on
the MNIST dataset\footnote{More details at \href{http://yann.lecun.com/exdb/mnist/index.html}{http://yann.lecun.com/exdb/mnist/index.html}.}.

\begin{figure}
  \centering
  \includegraphics[width=2.84413in,height=2.3631in]{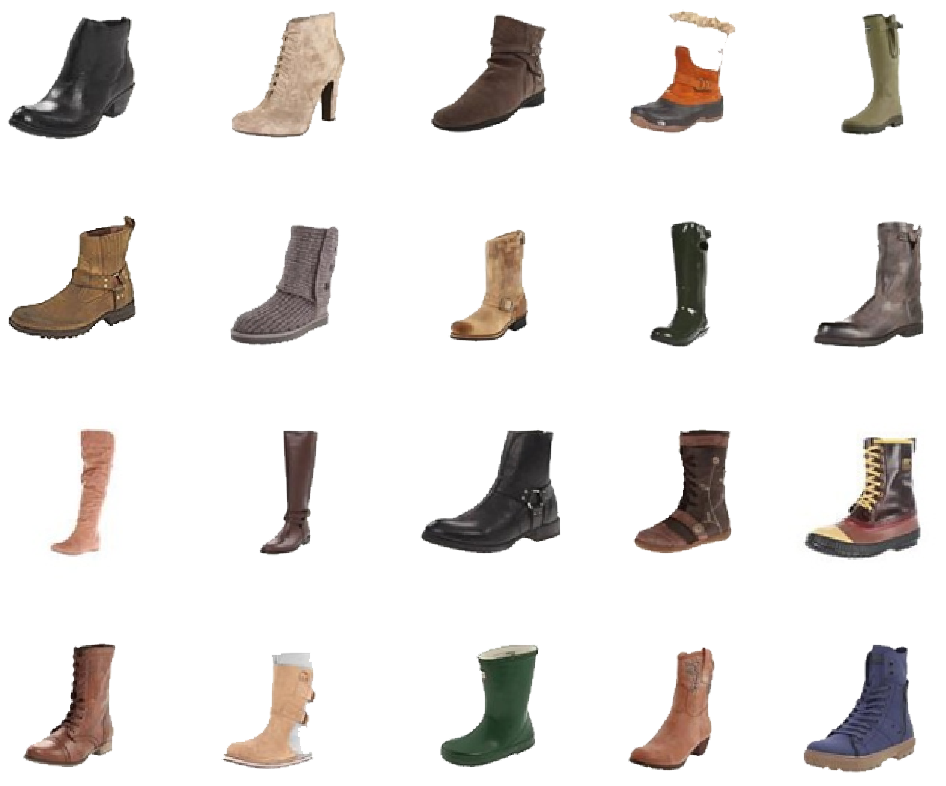}
  \caption{Examples of objects studied for similarity.}
  \label{Fig.1}
\end{figure}

\begingroup
\renewcommand{\section}[1]{\vspace{1em}\noindent\fontsize{12}{13}\textbf{#1}}
\section{2. Similarity evidence of segmentation}
\endgroup

\begingroup
\renewcommand{\subsection}[1]{\vspace{1em}\noindent\fontsize{11}{12}\textbf{#1}\vspace{1em}}
\subsection{2.1. Feature extraction and pruning}
\endgroup

Consider a trained FCN noted \(U\) with a set of selected layers
\(\phi\) consisting of \(N\) features, and an extraction procedure
applied to a one-class segmentation problem, where a single object is
segmented per image. To an object in \(\mathcal{X}\) corresponds a
predicted set of pixels (segment) generated by \(U\) in its last
convolution layer (output). Let \(f_{j}\) be an activated feature map of
\(\phi\), which constitutes the set \(F\ (\forall j \leq N\)). From this
map, there exists a set of pixels \(m_{i,j}\) responsive to an object
\(i\) from \(\mathcal{X}\) i.e., only a specific part of \(f_{j}\) is
activated with respect to an \(i\)th segment of an input image. The
activated pixels \(m_{i,j}\) are first extracted by cropping the segment
\(i\) from the last layer of \(U\), then propagating the cropped region
from segmentation to a specified \(f_{j}\) in \(\phi\) by re-adjusting
it with regards to the dimensions of \(f_{j}\) i.e., both, the centroid
position and size of the cropped segment region is adapted to the
resolution of lower-level feature maps. For instance, the propagation
from an output with size \(256 \times 256\), to a feature map of size
\(128 \times 128\) requires dividing the cropping size (and its centroid
coordinates) by 2. Note that due to spatial reduction and invariance,
the pixel-wise information of the original image object is altered in
the network after passing a filter over outputs with a given pixel
shifting (stride). Usually, convolutional layers pad the feature map if
the convolution crosses the output borders, which triggers a shift of
the activated signal. However, much of this signal (and position) can be
maintained within the propagating window using small and symmetric
kernels for convolution, pooling and up-sampling operations, and if
padding does not add up significant information on the edge of the
outputs, which is the case for the network used in this study (see
Section 3.1). \newline

Assuming \(M_{i} = \left\{ m_{i,1},\ldots,m_{i,N} \right\}\) a set of
extracted features derived from \(F\), a \(N\)-dimensional feature
vector \(v_{i}\) is first generated after describing every element of
\(M_{i}\) statistically. For example, the first statistical moment
(mean) of \(m_{i,j}\) is calculated, and can be regarded as an
activation magnitude noted \(x_{i,j}\). Accordingly, two objects with
same or very similar characteristics in \(\mathcal{X}\) presumably yield
analogous representations \(v_{i}\) in \(\mathcal{A}\) i.e.,
corresponding features \(m_{i,j}\) have similar magnitudes. This
assumption holds when objects from \(\mathcal{X}\) are of the same
dimension, ensuring that pattern comparisons via \(v_{i}\) are size
invariant. The data utilized in this research exhibits this
characteristic, as presented in Fig.1. Therefore, for each \(i\)th
object segmented by \(U\) from the image dataset UT Zappos50k (with size
\(s = |O|\)), the feature matrix denoting \(M\) can be represented as
\(X\). Each column \(u_{j}\) in \(X\) corresponds to a feature vector
with a probability density function (PDF) defined as \(p_{u_{j}}\).
Additionally, the row \(v_{i} = (x_{i,1},\ldots,x_{i,N})\) in \(X\)
corresponds to the feature vector or latent pattern of the \(i\)th
object. For similarity purpose, if we were to compare multiple patterns,
it may be convenient to reduce the dimensionality of \(X\). In this
study, sparse learning is employed to decrease the number of exploitable
(activated) features in \(\mathcal{A}\) during training (see method in
Appendix A). Subsequently, singular-value decomposition (SVD) is applied
to approximate the representation of \(X\), obtained after training, by
decomposing it into orthogonal and diagonal components, such that

\begin{equation}
\begin{aligned}
X_{s \times N} &= T_{s \times s}\Sigma_{s \times N}V_{N \times N}^{T}, \\ 
\end{aligned}
\end{equation}

where \(s = |O|\). \(T\) and \(V\) are orthonormal matrices comprising
singular vectors, which make up the columns of \(T\) and rows of
\(V^{T}\). \(\Sigma\) is a rectangular diagonal matrix whose diagonal
elements are positive singular values arranged in decreasing order.
Here, only a subset of relevant features can be determined by retaining
99\% of data variance from the activation magnitudes in \(X\). The first
(and highest) singular value in \(\Sigma\) is considered because its
first component accounts for the majority of the variances in \(X\).
Hence, the first singular vector of \(V^{T}\) is taken as a basis vector
to rank and withhold the \(n\) most relevant features (\(n < N\)). The
resulting network-related data matrix is noted as
\(X^{'} \in \mathbb{R}^{s \times n}\), with \(n\) being the number of
withheld activation magnitudes after pruning the \(N\) total network
features collected from \(\phi\). The \(i\)th mean activation vector is
defined as
\(v_{i} = \left( X_{i,j}^{'} \right)_{1 \leq j \leq n} = (x_{i,1},\ldots,x_{i,n})\),
and the components of the \(j\)th feature vector
\(u_{j} = \left( X_{i,j}^{'} \right)_{1 \leq i \leq s} = (x_{1,j},\ldots,x_{s,j})\)
are evaluated for similarity purpose.

\begingroup
\renewcommand{\subsection}[1]{\vspace{0.7em}\noindent\fontsize{11}{12}\textbf{#1}\vspace{0.7em}}
\subsection{2.2. Probabilistic transformation}
\endgroup

\begin{figure*}
  \centering
  \includegraphics[width=6.26772in,height=1.54724in]{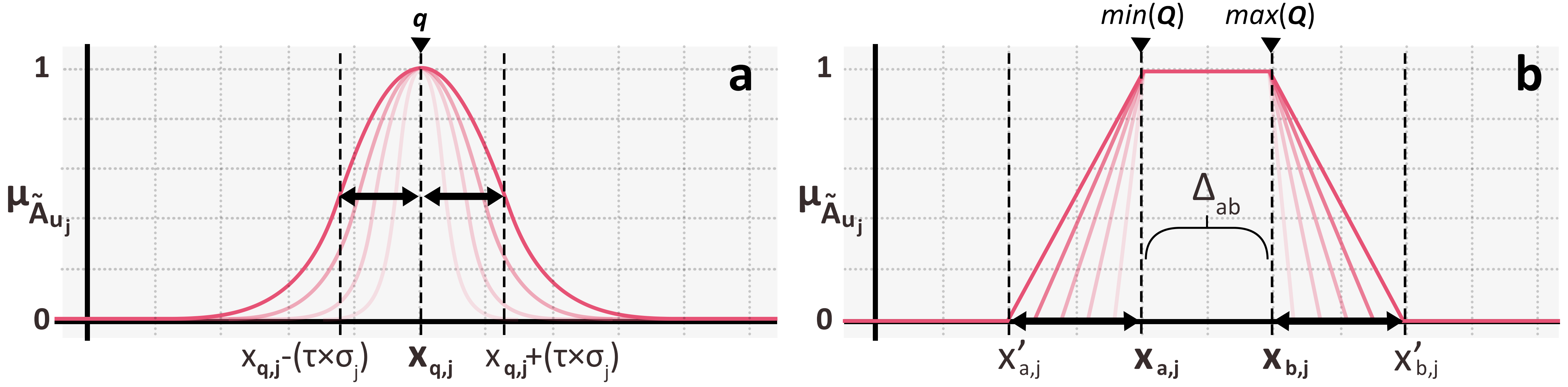}
  \caption{Membership functions: (a) Gaussian type, with \(x_{q,j}\) the central value of a \(j\)th feature given a query \(q\), and \(\sigma_j\) the standard deviation, and (b) trapezoidal, for a set of queries Q, with the lower and upper limits defined by \(x^{'}_{a,j}= x_{a,j}- \Delta_{ab}\), and \(x^{'}_{b,j}=x_{b,j}+\Delta_{ab}\), respectively. \(\Delta_{ab}\) is the range between the support limits \(x_{a,j}\) and \(x_{b,j}\), and \(\tau\in[0,1]\) is an arbitrary factor. Dark arrows indicate the varying uncertainty limits modulated by \(\tau\).}
  \label{Fig.2}
\end{figure*}

Consider two objects collected from \(O\) with similar characteristics
in \(\mathcal{X}\): an object indexed \(q\) that we wish to study (a
query) and an object indexed \(i\) selected for comparison. Presumably,
their feature patterns may be analogous i.e., the activation magnitudes
\(x_{i,j}\) of related feature vectors (\(v_{i}\) and \(v_{q}\)) shall
be close. Given this assumption, we can denote a confidence interval of
values possibly determining \(v_{q}\), centered at a point estimate
\(x_{q,j}\) in \(p_{u_{j}}\). From a probabilistic perspective, this
confidence interval is fuzzy, meaning that as adjacent values approach
\(x_{q,j}\), the probability of determining \(v_{q}\) increases. This
fuzziness can be measured using a membership function
\(\mu_{{\widetilde{A}}_{u_{j}}\ }\), which transforms \(u_{j}\) into a
fuzzy set \({\widetilde{A}}_{u_{j}}\). The choice of
\(\mu_{{\widetilde{A}}_{u_{j}}}\) depends on the a priori knowledge of
data from the practitioner, or the type of object query desired. For
instance, utilizing a Gaussian membership function enables obtaining
\({\widetilde{A}}_{u_{j}}\) using the equation

\begin{equation}
\begin{aligned}
\mu_{{\widetilde{A}}_{u_{j}}}^{q}\left( x_{i,j} \right) = e^{- \left( \frac{x_{i,j} - x_{q,j}}{\tau\sigma_{j}} \right)^{2}}, \\ 
\end{aligned}
\end{equation}

\begin{figure*}
  \centering
  \includegraphics[width=6.72569in,height=2.47014in]{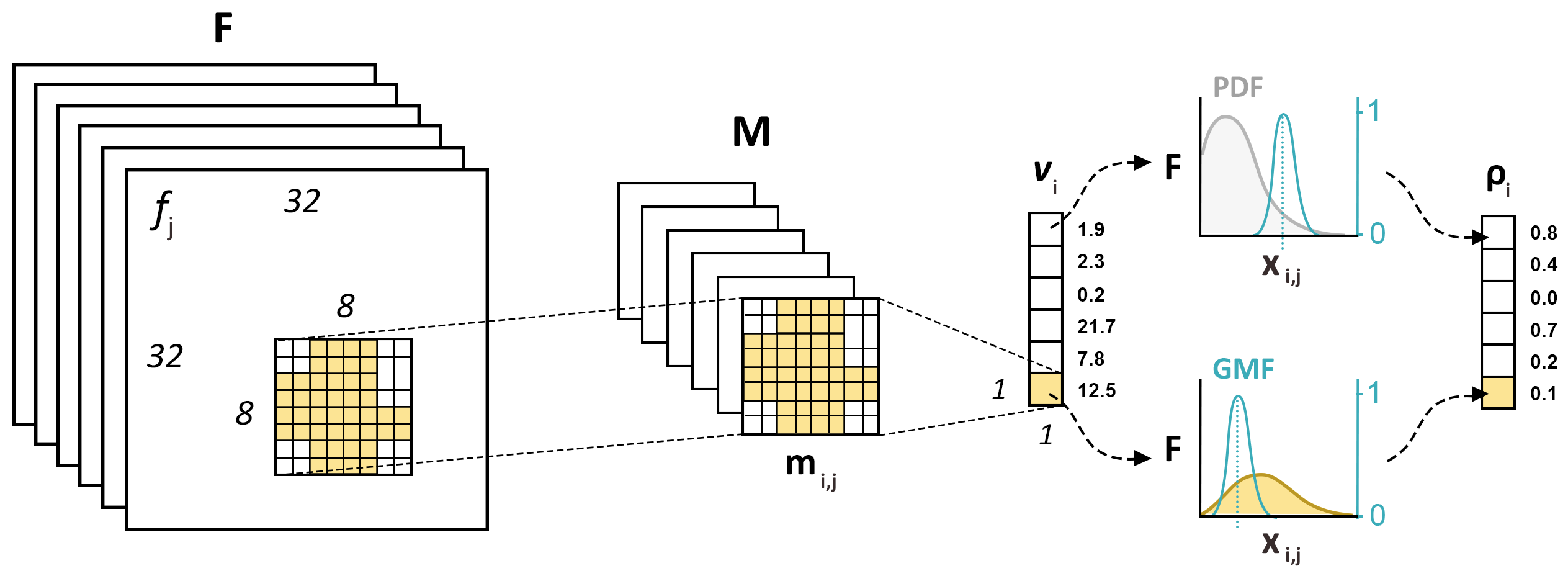}
  \caption{Object-oriented extraction of feature \(m_{i,j}\) from an activated feature \(f_j\). Resulting set \(M\) is converted into an activation vector \(v_i\) by statistical description of the activated region of \(m_{i,j}\) (yellow mask). Each component \(x_{i,j}\in v_i\), associated to a feature \(f_j\), defines a probability density function (PDF) computed for a collection of objects (\(\forall i\in O\)). A sampled value of this PDF, specified by a query \(q\), corresponds to the centroid (mean) of a Gaussian membership function (GMF), which maps the PDF crisp values into the range [0,1]. Each \(x_{i,j}\) is thus converted into a fuzzy value as a function of its distance to the centroid. Resulting vector \(\rho_i\) is interpreted as a feature-wise probabilistic similarity measurement of an \(i\)th object, with regards to the queried one indexed \(q\).}
  \label{Fig.3}
\end{figure*}

with \(x_{q,j}\) being the central value defined by a query \(q\), and
\(\sigma_{j}\) the standard deviation of \(u_{j}\) defining the
confidence interval scaled by \(\tau\). The factor \(\tau\) modulates
the spread of the distribution (Fig.2a), influencing the precision of
probabilistic estimates. It is a relevant property to decide the degree
of confidence in probability search, as decided by the practitioner.
Note that \(u_{j}\) is normalized prior to applying fuzzy transforms.
Using equation (2), the fuzzified elements of \(u_{j}\) constitute fuzzy
numbers of a pattern set to analyze. Considering that
\(\mu_{{\widetilde{A}}_{u_{j}}\ }:u_{j} \rightarrow {\lbrack 0,1\rbrack}^{s}\),
we denote
\(g_{i,j}^{q} = \mu_{{\widetilde{A}}_{u_{j}}}^{q}\left( x_{i,j} \right)\)
as the grade of the membership of an element \(x_{i,j}\) in
\({\widetilde{A}}_{u_{j}}\) with respect to an object \(q\), which
yields a probability value within the interval {[}0,1{]}. By doing so
for a feature vector \(v_{i}\), its probabilistic counterpart
\(\rho_{i}^{q} = \left( g_{i,1}^{q},\ldots,g_{i,n}^{q} \right)\) can be
obtained. \newline

For situations where the practitioner attempts examining multiple
queries in parallel (\(q_{1},q_{2},q_{3}\) and so on), it may be
convenient to use e.g., a four-sided trapezoidal membership function to
approximate \(\rho_{i}^{Q}\), with \(Q\) being a set of queries
(Fig.2b). This function is written

\begin{equation}
\begin{aligned}
\mu_{{\widetilde{A}}_{u_{j}}}^{Q}\left( x_{i,j} \right) = \left\{ \begin{array}{r}
\ \  \\
1,\ \ x_{a,j} \leq x_{i,j} \leq x_{b,j} \\
\begin{matrix}
\frac{x_{i,j} - x_{a,j}^{'}}{x_{a,j} - x_{a,j}^{'}},\  & x_{a,j}^{'} \leq x_{i,j} \leq 
\end{matrix}\ x_{a,j} \\
\ \  \\
\begin{matrix}
\frac{x_{b,j}^{'} - x_{i,j}}{x_{b,j}^{'} - x_{b,j}},\ \  & x_{b,j} \leq x_{i,j} \leq x_{b,j}^{'}
\end{matrix} \\
0,\ \ elsewhere\ 
\end{array} \right.\ , \\ 
\end{aligned}
\end{equation}

with \(x_{a,j}\) and \(x_{b,j}\) being the lower and upper support
limits determined by \(min(x_{q_{1},j},{\ldots,x}_{q_{i},j})\) and
\(max(x_{q_{1},j},{\ldots,x}_{q_{i},j})\), respectively. \(x_{a,j}^{'}\)
and \(x_{b,j}^{'}\) are the lower and upper limits of
\(\mu_{{\widetilde{A}}_{u_{j}}}^{Q}\) defined by
\(x_{a,j}^{'} = \ x_{a,j} - \ \mathrm{\Delta}_{ab}\), and
\(x_{b,j}^{'} = x_{b,j} + \mathrm{\Delta}_{ab}\) respectively. The term
\(\mathrm{\Delta}_{ab}\) defines the ratio of the range
\(\lbrack x_{a,j},x_{b,j}\rbrack\) written
\(\mathrm{\Delta}_{ab} = \ \tau \times (x_{b,j} - x_{a,j})\), and
\(\tau \in \lbrack 0,1\rbrack\) is an arbitrary factor modulating the
lower and upper limits. Using a trapezoidal function, the confidence
interval of the \(j\)th feature is determined by comparing \(x_{i,j}\)
for each \(i\)th object queried in \(Q\). In contrast, the Gaussian
function utilizes the standard deviation of \(p_{u_{j}}\). Note that
both, the Gaussian and trapezoidal functions are scaled by \(\tau\),
which is experimentally evaluated. An overview of the feature extraction
procedure, followed by transformation and similarity analysis is shown
in Fig.3.

\renewcommand{\tablename}{TABLE}
\renewcommand{\thetable}{I}
\begin{table*}[t]
\centering
\caption{Simplified U-Net architecture. Layers have two sequential \(3\times 3\) convolutions (e.g. C11 and C12, with 32 filters per layer), followed by max-pooling (encoder) or preceded by concatenation (decoder). ReLU is the activation function. Output size is given in pixels (side-by-side). N-out: network output (\(1\times 1\) convolution, softmax); BN: bottleneck. Selected layers for similarity analysis comprise those of the encoder, BN, and the decoder, which represents 1,600 features in total.}
\begin{tabularx}{0.8\textwidth}{>{\raggedright\arraybackslash}p{0.15\textwidth}XXXXXXXXXXX}
\toprule
\textbf{U-Net} & \multicolumn{4}{c}{\textbf{Encoder}} & \textbf{BN} & \multicolumn{4}{c}{\textbf{Decoder}} & \textbf{N-out} \\
\cmidrule(lr){2-5} \cmidrule(lr){7-10}
\textbf{Layers} & C11, C12 & C21, C22 & C31, C32 & C41, C42 & C51, C52 & C61, C62 & C71, C72 & C81, C82 & C91, C92 & C101, C102 \\
\midrule
\textbf{Output size} (pixels) & 256 & 128 & 64 & 32 & 32 & 32 & 64 & 128 & 256 & 256 \\
\textbf{Total filters} & 64 & 128 & 192 & 256 & 320 & 256 & 192 & 128 & 64 & 2 \\
\bottomrule
\end{tabularx}
\end{table*}

\begingroup
\renewcommand{\subsection}[1]{\vspace{0.7em}\noindent\fontsize{11}{12}\textbf{#1}\vspace{0.7em}}
\subsection{2.3. Probabilistic similarity}
\endgroup

To an object pattern \(v_{i}\) corresponds an encoded counterpart
\(\rho_{i}^{q}\) with respect to a queried pattern \(v_{q}\).
Considering equation (2), a fixed parameter \(\tau\), and \(n\)
non-pruned features, the probabilistic similarity of an \(i\)th object
with respect to a queried one indexed \(q\) is given by the formula

\begin{equation}
\begin{aligned}
P\left( v_{i}\text{|}v_{q} \right) = \frac{1}{n}\sum_{j = 1}^{n}{\omega_{j} \times g_{i,j}^{q}}, \\ 
\end{aligned}
\end{equation}

where
\(P\left( v_{i}\text{|}v_{q} \right):\mathbb{R}^{n}\mathbb{\longrightarrow R \in}\lbrack 0,1\rbrack\),
and \(\omega_{j}\) defines the feature relevance in classifying objects,
such that, \(\omega_{j} \geq 0,\ \forall j \leq n\) and
\(\sum_{j = 1}^{n}\omega_{j} = 1\). Here, \(\omega_{j}\) constitutes a
weight vector \(\Omega\). Presumably, the \(n\) network features do not
have equal contribution in classification i.e., some features \(u_{j}\)
better distinguish types of objects when conducting a similarity search.
Hence, the similarity measure can be essentially determined by an
unknown subset of network features. To the extent where contributions
are equal (\(\omega_{j} = \frac{1}{n}\)), the associated probabilistic
similarity may be less precise. The relevance of this contribution per
feature can defined by weights estimated from the SVD method (see
Section 2.1), or calculated from the summed difference of mean
activations between similar objects. To exemplify the later, assume a
set of clusters \(C\), where each cluster \(C_{q}\) designates several
objects with similar feature patterns. Here, the subscript \(q\) relates
to a query i.e., an object index making a set noted \(\Theta\), and it
does not serve as iteration index. \(C_{q}\) can be defined by the
practitioner by inspecting objects in \(\mathcal{X}\) visually with the
help of equation (4) when conducting a similarity analysis. By doing so, a pool of a given number of analogous objects obtained for a
query \(q\) via equation (4) is considered being a cluster \(C_{q}\).
Next, the mean activation of every cluster is determined component-wise
e.g., given \(C_{1}\), we denote
\(v_{C_{1}} = ({\overline{x}}_{C_{1},1},\ldots,{\overline{x}}_{C_{1},n})\)
where
\({\overline{x}}_{C_{1},\ 1} = \frac{1}{\left| C_{1} \right|}\sum_{i \in C_{1}}^{}x_{i,1}\).
Finally, the differences of mean activations between clusters
(\(v_{C_{1}}\), \(v_{C_{2}}\), \(v_{C_{3}}\), and so on) is calculated
and summed component-wise, such that

\begin{align}
\omega_{j} &= \sum_{\substack{(i,\ k) \in \Theta, \\ i \neq k}} \left| {\overline{x}}_{C_{i},j} - {\overline{x}}_{C_{k},j} \right|. \label{eq:omega}
\end{align}

The resulting sum determines the weight vector \(\Omega\), which is
thereafter normalized. Note that equation (5) can be applied to an
object pattern \(v_{i}\) (instead of \(v_{C_{q}}\)) if no initial pool
of objects (or clusters) is defined.

\section{3. Experiments}
\subsection{3.1. Implementation details}

A U-Net model with a standard architecture (Table 1) {[}16{]} is trained
to segment single-object images from the UT Zappos50k dataset. The
12,833 RGB images constituting the dataset have a fixed size of
\(256 \times 256\) pixels individually, and the represented objects
(boots) have distinguishable colors, textures, shapes and details. To
improve the model robustness, data augmentation for shift, rotation and
scaling invariance are applied. The energy function is computed by a
pixel-wise softmax over the final feature map combined with the
cross-entropy loss function. The initial weights are sampled from a Gaussian distribution with Xavier weight initialization, while Adam {[}19{]} was employed for optimizing the training procedure with an
initial learning rate of 1e-3.

\renewcommand{\tablename}{TABLE}
\renewcommand{\thetable}{II}

\begin{table*} 
\centering
\caption{Model performance with varying sparsity ratios per convolutional layer (\(R^{0}_{sp}\)). OE: optimal epoch. \(DC= 2\times\frac{Precision \times Recall}{Precision + Recall + \epsilon}\) , with \(\epsilon \) being a constant value to avoid division by 0.}  

\begin{tabularx}{0.6\textwidth}{%
  >{\raggedright\arraybackslash}p{0.08\textwidth}XXXXXX
}
\toprule\noalign{}
\(\mathbf{R}_{\mathbf{sp}}^{\mathbf{0}}\) & \textbf{OE} & \textbf{Training} & \textbf{Validation} & \textbf{Training} & \textbf{Validation} \\
\midrule\noalign{}
None & 531 & 0.017 & 0.013 & 0.982 & 0.991 \\
0.3 (30\%) & 1237 & 0.031 & 0.028 & 0.979 & 0.980 \\
0.5 & 1385 & 0.033 & 0.036 & 0.966 & 0.975 \\
0.7 & 1906 & 0.034 & 0.039 & 0.969 & 0.971 \\
\hline
\end{tabularx}
\end{table*}

\renewcommand{\tablename}{TABLE}
\renewcommand{\thetable}{III}
\begin{table*}[t]
\centering
\caption{Residual non-zero features of convolutional layers considering \(R^{0}_{sp}\) and SVD.}
\begin{tabularx}{0.8\textwidth}{
  >{\raggedright\arraybackslash}p{0.18\textwidth}XXXXXXXXXXX
}
\toprule
\textbf{Layers} & C11, C12 & C21, C22 & C31, C32 & C41, C42 & C51, C52 & C61, C62 & C71, C72 & C81, C82 & C91, C92 & \textbf{Total} \\
\midrule
\textbf{Initial} & 64 & 128 & 192 & 256 & 320 & 256 & 192 & 128 & 64 & 1600 \\
\(\mathbf{R}_{\mathbf{sp}}^{\mathbf{0}}\) (70\%) & 35 & 60 & 61 & 58 & 34 & 71 & 96 & 69 & 38 & 532 \\
\(\mathbf{R}_{\mathbf{sp}}^{\mathbf{0}}\) (70\%) \textbf{+} \textbf{SVD} (99\%) & 26 & 28 & 57 & 54 & 32 & 65 & 96 & 68 & 27 & 453 \\
\bottomrule
\end{tabularx}
\end{table*}

\subsection{3.2. Dimensionality reduction}

Training experiments with different sparsity levels were conducted (see
Table 2). Sparse learning was done with parameters
\(\alpha \in \lbrack 0.1,1\rbrack\) and \(\lambda = 1\) (see Appendix
A), considering a model with a maximum level of sparsity at 70\% i.e.,
with 30\% non-zero activated features which can be analyzed from
\(\mathcal{A}\).

Above this level, training can be incredibly longer. As shown in Table
2, the model achieved a loss of 0.034 (0.039 for validation) with
pixel-level accuracy (F1 score) of 0.969 (0.971), considering a sparsity
level of 70\%. Sparse learning excluded, the model exhibits better
performance, with a loss and accuracy of 0.017 and 0.982, respectively.
Nevertheless, results demonstrate that, although there is an apparent
loss in performance with sparsity, this performance does not drop
significantly with increasing sparsity. Note that due to the nature of
the dataset used in this study, a single-class problem per image was
addressed, and objects are clearly separable from a blank (white)
background. As such, a performance drop for datasets with higher data
variance may be expected.

After constituting the data matrix \(X\) from \(M\) using a 70\% sparse
model, a feature pruning via SVD is applied to obtain \(X^{'}\) whose
features retain 99\% of the data variance. Table 3 presents residual
features after sparse learning and pruning. 453 features are utilized
for similarity evidence out of 1,600 from the selected layers \(\phi\)
(Table 1).

\subsection{3.3. Weighting scheme}

Assuming that the 453 examined features may contain class-specific
information, their importance is approximated for weighting purposes. To
accomplish this, an initial cluster is visually determined from the
data, aided by similarity searches estimated with equation (4) for
multiple single queries. Subsequently, the mean feature vector of each
cluster is determined from the normalized matrix \(X^{'}\). Finally, the
differences between these vectors are summed using equation (5) and
normalized to obtain the resulting magnitude change.

As shown in Fig.4, these magnitude changes confirm that the model's
discriminative performance is not equally shared among the 453 features.
From a statistical standpoint, many of these features moderately
contribute in classifying (\textasciitilde63\% achieves magnitudes
between 0.1 and 0.6), some highly contribute to it (23\% above 0.6), and
a minor portion has low activations (close to 0) for given clusters (up
to \textasciitilde14\%). Interestingly, none of the convolutional layers
seem to play a dominant role in classifying, although the distribution
of total magnitude change is not even between these layers (e.g., 7\%
and 22\% in C3x and C6x, respectively). Notably, there is an apparent
difference on magnitude variability between the encoder (27\% of change
in the network), which extracts contextual information and transferable
patterns, and the decoding part of U-Net (53\%), which learns how to
apply these patterns while mapping the latent features to the data
space. This suggests that the decoder may also have significant
influence on data speciation. It is common for lower layer features to
generate general-purpose representations, which are not class agnostic. The same is true for higher level features when combining lower representations. In this study, representation learning (and feature speciation) has benefited from parameters recalibration, induced by channel-wise sparse learning.

\begin{figure*}
  \centering
  \includegraphics[width=5.68472in,height=3.35972in]{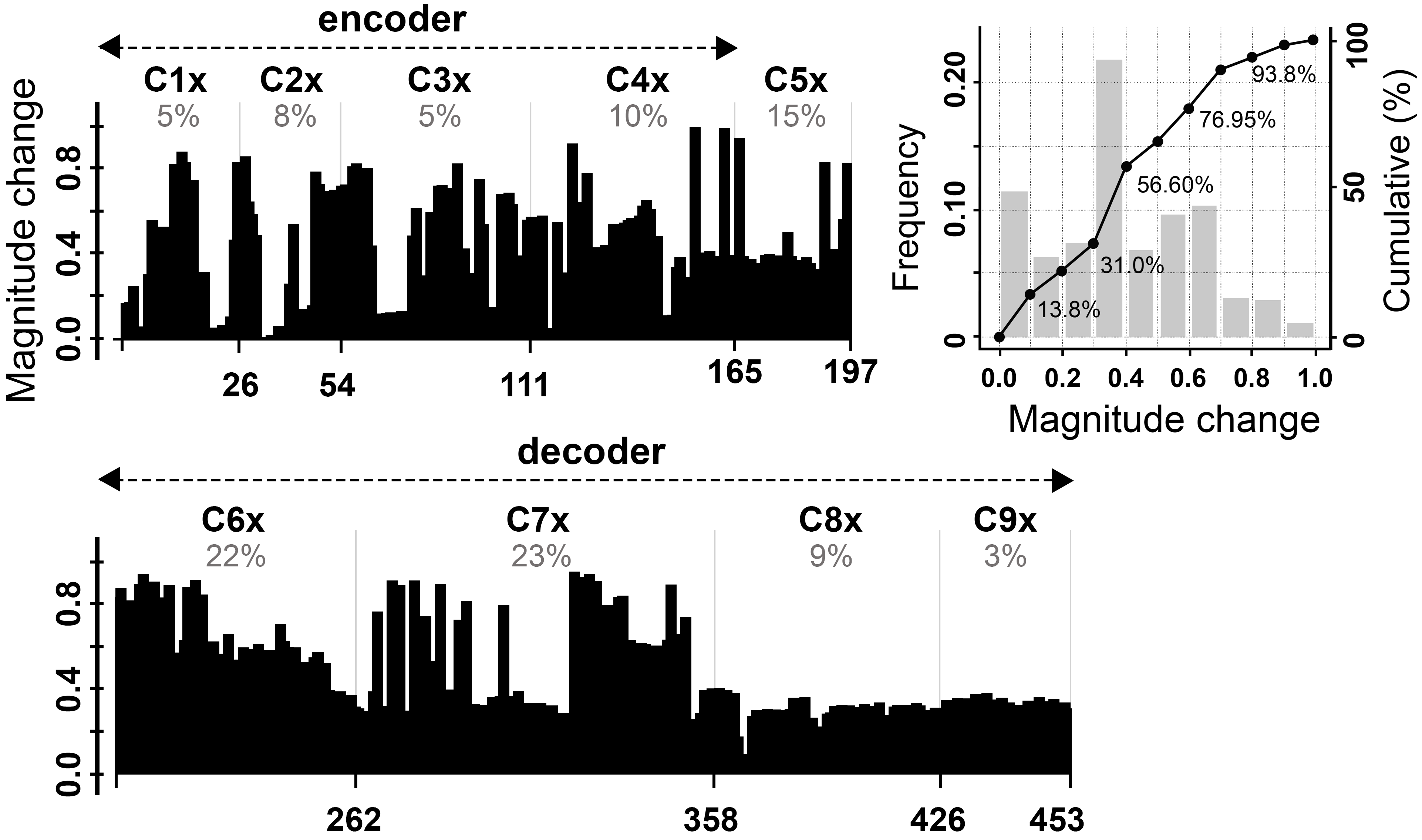}
  \caption{Normalized sum of mean activation changes calculated between 12 clusters, each having a minimum of 20 items. Percentage of total magnitude change is indicated per pairs of convolutional layers (e.g., C11 and C12 defined as C1x, see Table 1). Related distribution is shown in both, frequency and cumulative forms.}
\end{figure*}

\subsection{3.4. Similarity analysis}

Preliminary results for similarity analysis help identifying data
patterns among the 12,833 boots studied. Especially, queried objects
appear to share characteristics with other similar instances in terms of
colorimetry, texture and shape (Fig.5). However, this similarity
decreases if such characteristics differ in the pool of analyzed objects
i.e., the similarity score is less than 0.6 (see the last query example
in Fig.5), or when investigating multiple queries jointly (Fig.6).
Besides this, exploiting a large fraction of convolutional layers in
similarity analysis seems important for satisfying results, given the
speciation performance of these layers, as described in Section 3.3 and
Fig.4. It should be noted that conducting the similarity analysis solely
on individual parts of the network, such as the decoder or encoder
alone, did not yield satisfactory results. Nonetheless, analogous
results can still be achieved using features extracted from a ResNet50
model, trained to classify e.g., digits from the MNIST dataset (see
Appendix B). This demonstrates the feasibility of applying the
analytical approach for similarity computations described in this
research to non-FCN models.

\begin{figure*}
  \centering
  \includegraphics[width=5.22708in,height=5.2375in]{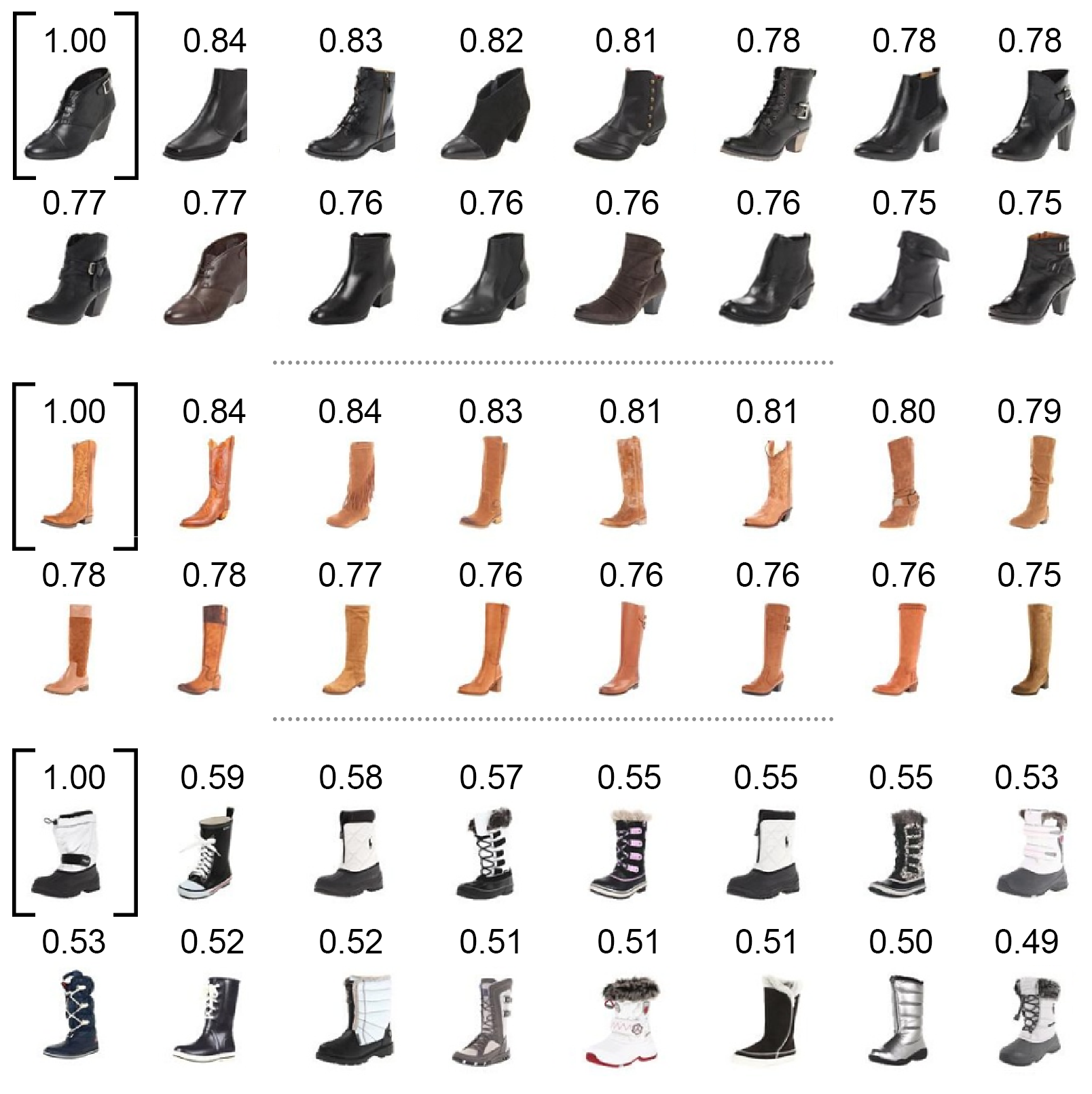}
  \caption{Examples of probabilistic similarity results for single queries (indicated by black square brackets) using a Gaussian membership function. Average values are indicated in decreasing order from top left to bottom right. Only 16 objects are shown per search.}
  \label{Fig.5}
\end{figure*}

\begin{figure*}
  \centering
  \includegraphics[width=4.80417in,height=2.43611in]{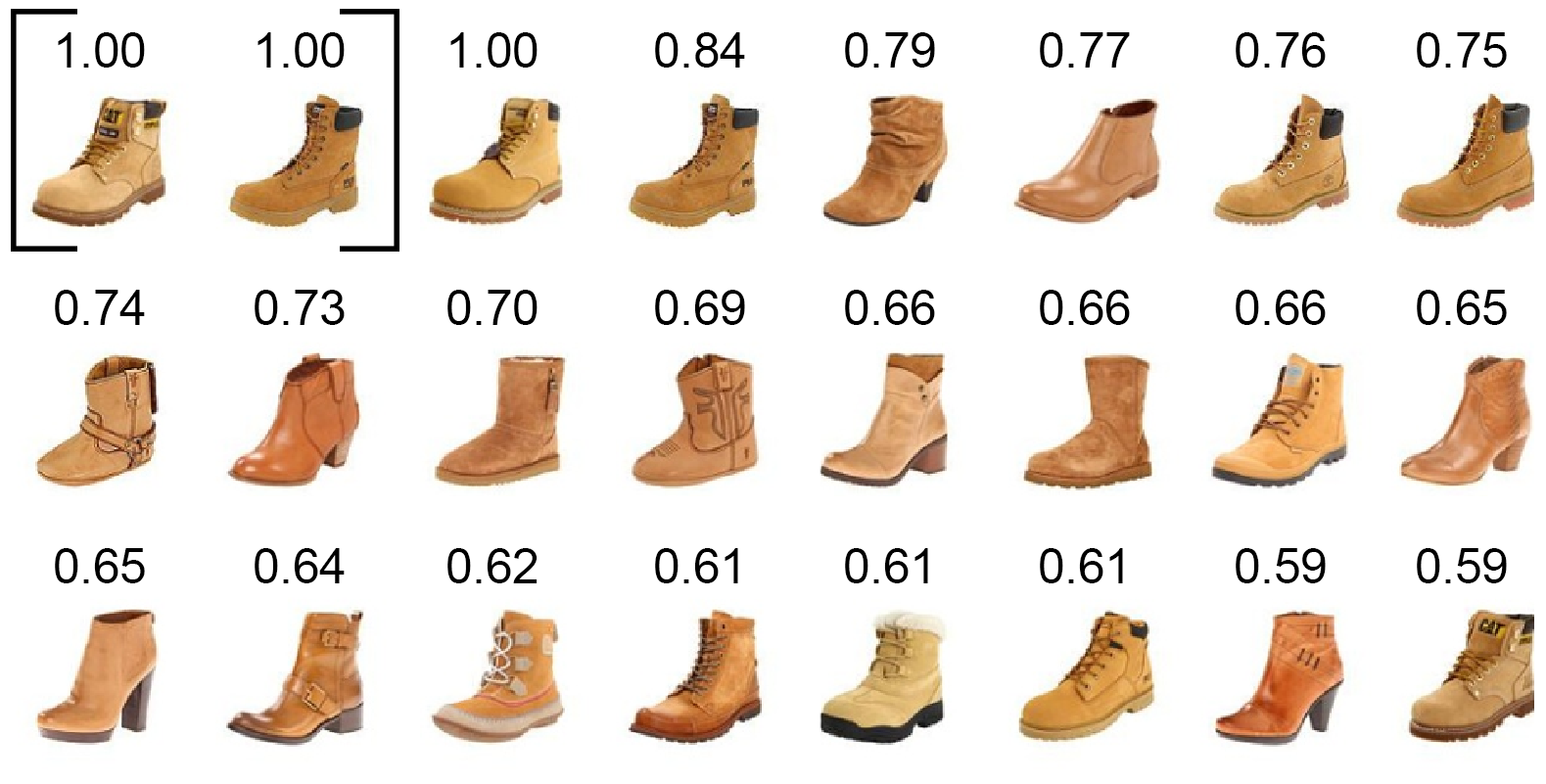}
  \caption{Example of probabilistic similarity results for two queries (indicated by black square brackets) using a trapezoidal membership function. Average values are presented in decreasing order from top left to bottom right. Only 24 objects shown.}
  \label{Fig.6}
\end{figure*}

\section{4. Discussion}

In this study, the distinctiveness of objects is objectively determined
by the similarity algorithm and features of a trained network. It is
more likely that object instances with consistent patterns i.e., well
represented in data, end up having more similar objects with higher
probabilistic values (closer to 1). Furthermore, the similarity search
strategy is made possible with membership functions where subsets of
activation magnitudes falling within the neighborhood of a query vector
\(v_{q}\) are grouped. Further studies may explore different search
functions to construct fuzzy queries, such as triangular functions,
without changing the properties of the method described in Section 2.
More generally, it seems important to adapt similarity measures with the
type of data analyzed. Notably, to the extent that subsets of activation
patterns are evaluated by Gaussian functions, the shape of the search
function is adjustable with a factor \(\tau\), as defined in equation
(2). Necessarily, a large \(\tau\) implies to search for similar
elements with lower memberships, which may be relevant for image
datasets whose 2D objects have high variations in appearance. \newline

The analysis of similar patterns in the UT Zappos50k dataset allows us
to distinguish key recurring characteristics among different types of
boots. These characteristics are embedded in the latent features of
trained networks, and extracting them as fuzzy values helps determining
valuable information from data. Qualitative evaluations of similar
objects depicted in Fig. 5 and 6 offer interpretative insights,
highlighting that certain boot details like colour, texture, and shape
may carry more significance at the feature level, than minor
characteristics like laces, buckles, and heels. Consequently, an object
query does not inherently match perfectly other objects due to
variations in elements such as laces, buckles, or heels. To achieve
better matching, the extracted network information pertaining to these
spatial elements should be assigned greater importance in the analysis.
To do so, future practitioners may want to extract and evaluate latent
features specific to certain object details, alongside with the main
object\textquotesingle s latent features. The work of {[}15{]}
demonstrates this approach by dissecting both, the main object class,
and its peculiar details as subclasses, giving extra importance to
subclasses in similarity calculations by including latent information of
each subclass in the analysis.

\section{5. Conclusion}

In this paper, a novel object-oriented similarity analysis method with
fuzzy evidence of transformed network features facilitates the
distinction of key patterns in data. The proposed approach demonstrates
successful performance in image segmentation and effectively identifies
object candidates based on attributes from one or multiple queries. The
combination of transforming features from pre-trained models, fuzzy
inference, and sparse learning allows for an intuitive determination of
possible groups of similar 2D objects (surrogate classes) sharing
analogous characteristics. This approach could be iteratively applied
for clustering purposes, possibly by using a sequential partitioning of
probabilistic similarity values. In addition, it can be envisioned to
leverage the concept of fuzzy neighborhood in discretized feature space.
Although sparse learning and pruning techniques are employed to reduce
the dimensionality of \(\mathcal{A}\), a significant portion of it
serves as the foundation for similarity search. As such, it may be
relevant to investigate alternative network architectures specifically
tailored to the similarity scheme of this study to improve similarity
results.

\section{Appendix A: Sparse learning}

Several studies utilize sparsification techniques to mitigate
overparameterization and overfitting in neural networks. For further
details, interested readers may refer to the comprehensive review by
{[}20{]}. Sparsity often refers to the property of model parameters to
contain numerous zero coefficients (sparsity of parameters), which may
be exploited for computational and resource savings {[}21, 22, 23{]} or
model simplification {[}24, 25{]}. When combined with pruning, sparse
models can achieve high compression through sparse connectivity {[}26,
27, 28{]}. However, it is common to trade off accuracy for the reduction
in model size and efficiency {[}29, 30{]}. In this study, the objective
is not to compress networks or accelerate learning. Instead, the focus
is on reducing the dimensionality of the activation space
\(\mathcal{A}\) during training, thereby pruning inactive (or
unimportant) neural representations post-training for subsequent
similarity evidence. Hence, the utilization of loss functions for both
the main task and sparse learning aids in reducing active features
during training on a selection of layers \(\phi\). In this context,
sparsity denotes the ratio of zeroed-out layer outputs at the channel
level, which are pruned after completing the training process. Pruning
is achieved via singular-value decomposition of a data matrix derived
from \(\mathcal{A}\) (see Section 2.1). It is important to note that the
level of sparsity and pruning carries significant implications: (i) it
simplifies and reduces the turnaround time of post-training feature
extraction and examination using multivariate methods, (ii) it mitigates
the high-dimensional characteristics of \(\mathcal{A}\) (e.g., the curse
of dimensionality), particularly when the number of network features
studied from \(\phi\) surpasses the number of observations (e.g., in the
case of deep models trained on small datasets), and (iii) it promotes
model generalization. Maximizing feature speciation in \(\mathcal{A}\)
is crucial, ensuring that non-pruned features are more likely to
distinguish themselves through neural specialization, particularly in
tasks like classification. Accordingly, some features are more relevant
when studying activation magnitudes of \(v_{i}\) for similarity
analysis. This speciation is particularly of interest to induce
heterogenous feature weighting (as formulated in Section 2.3).

To address sparse learning, penalties are applied using task- and
sparse-related loss functions. Mathematical notations provided below are
independent to notations used outside the Appendix A in this paper. Let
\(D\) be a dataset of \(P\) input-output pairs
\({\left\{ \left( z_{i},y_{i} \right) \right\}\ }_{i = 1}^{P}\), where
\(z_{i}\) is an input sample and \(y_{i}\) the corresponding annotation.
Assume \(\Phi( \bullet ;\theta)\) being the learning model parametrized
by the weights \(\theta\), which takes the input-output pairs of \(D\)
and optimizes \(\theta\) given the cross-entropy loss
\(\mathcal{L( \bullet )}\). In the following, biases are omitted for
notational simplicity. The task loss can be set by the formula

\begin{equation}
\begin{aligned}
\mathcal{L}_{task} = \frac{1}{P}\left( \sum_{i = 1}^{P}{\mathcal{L}\left( \Phi\left( z_{i};\theta \right),y_{i} \right)} \right).
\end{aligned}
\tag*{(A1)}
\end{equation}

A convolutional layer \(l\) transforms an input tensor
\(z^{l} \in \mathbb{R}^{H \times W \times C}\) into an output tensor
\(y^{l} \in \mathbb{R}^{H^{'} \times W^{'} \times K}\) using the filters
\(\theta^{l} \in \mathbb{R}^{K_{s} \times K_{s} \times C \times K}\),
given \(C\) inputs and \(K\) output features. \(H\) and \(W\) represent
respectively the height and width of inputs, while \(K_{s}\) is the
kernel size. Input tensors are individually convoluted with a set of 2D
kernels \(\left\{ \theta_{j,k}^{l} \right\}_{k = 1}^{C}\) such that the
\(j\)th ouput \(y_{j}^{l}\) is the sum of resulting convolutions
(\(\forall j \leq K\)). The output is computed by the formula

\begin{equation}
\begin{aligned}
y_{j}^{l} = \sum_{k = 1}^{C}{\theta_{j,k}^{l}*z_{k}^{l}},\ \ s.t.\ \ y_{j}^{l} \in \mathbb{R}^{H^{'} \times W^{'}}.
\end{aligned}
\tag*{(A2)}
\end{equation}

Let \(u_{j}^{l}\) be the activated counterpart of \(y_{j}^{l}\) using
ReLU as the activation function. The channel-level ratio of
non-zero (active) features \(R_{sp} \in \lbrack 0,1\rbrack\) is calculated for a
selection of layers \(\phi\) by converting \(u_{j}^{l}\) into its
respective scalar \(r_{j}^{l}\):

\begin{equation}
\begin{aligned}
R_{sp} = \frac{1}{|\phi| \times K}\sum_{l \in \phi}^{}{\sum_{j = 1}^{K}{sgn\left( r_{j}^{l} \right)}},
\end{aligned}
\tag*{(A3)}
\end{equation}

where
\(\ r_{j}^{l} = \sum_{k,q}^{}u_{j,k,q}^{l},\ \forall k \leq H^{'},\ \forall q \leq W^{'}\).
The sign function \(sgn( \bullet )\) converts \(r_{j}^{l}\) to 1 if it
is non-zero (positive), 0 otherwise. The opposite counterpart of
\(R_{sp}\) is the sparsity ratio \(R_{sp}^{0} = 1 - R_{sp}\). To
increase \(R_{sp}^{0}\), negative values of \(\theta^{l}\) are truncated
to zero such that \(\theta_{+}^{l} = max\left( 0,\ \theta^{l} \right)\),
and penalize the positive ones in \(\theta_{+}^{l}\). To do so,
\(\theta_{+}^{l}\) is redefined as a \(K\)-dimensional vector
\(\xi^{l}\) by summing its elements from dimensions \(K_{s}\) and \(C\)
over \(K\). This vector is then summed with the \(K\)-dimensional bias
vector of layer \(l\) to constitute the vector
\(\xi^{l} = (e_{1,}^{l},\ldots,e_{K}^{l})\). The symbol \(\xi\) denotes
the resulting sum for notational convenience. The loss for channel-wise
sparsity is given by

\begin{equation}
\begin{aligned}
\mathcal{L}_{sp} = \frac{1}{|\phi| \times K}\sum_{l \in \phi}^{}{\sum_{j = 1}^{K}e_{j}^{l}}.
\end{aligned}
\tag*{(A4)}
\end{equation}

Minimizing \(\mathcal{L}_{sp}\) is trivial compared to the task loss
\(\mathcal{L}_{task}\), although reaching an objective \(R_{sp}^{0}\)
may sometimes require to re-scale \(\mathcal{L}_{sp}\) in the final
objective function. A weight factor \(\lambda \in \mathbb{R}_{0}^{+}\)
can be used for this purpose in the regularized objective function
\(\mathcal{R}\):

\begin{equation}
\begin{gathered}
\begin{aligned}
\mathcal{R}(\theta) &= \mathcal{L}_{\text{task}} + \lambda\mathcal{L}_{\text{sp}} \times \gamma, \\
\theta^{*} &= \underset{\theta}{\text{arg min}}\left\{ \mathcal{R}(\theta) \right\},
\end{aligned}
\end{gathered}
\quad
\gamma = \left( \frac{R_{\text{sp}} - \beta}{1 - \beta} \right)^{\alpha}.
\tag*{(A5)}
\end{equation}

The term \(\gamma\) limits sparse learning at a target non-zero features
objective \(\beta \in \lbrack 0,1\rbrack\), given a learning speed
defined by \(\alpha\). The variables \(\lambda\) and \(\alpha\) are both
evaluated experimentally.

\section{Appendix B: Similarity analysis of digits from the MNIST dataset}
A ResNet50 model is trained on the MNIST dataset with a sparsity level of 50\%. This sparsity level is chosen to strike a balance between preserving essential features and reducing redundancy. Once the model is trained, similarity computations is conducted using the analytical procedure described in Section 2. In contrast to the conventional approach, where only a section of the feature map guided by segmentation outputs is used for magnitude calculations, the entire feature map generated by convolution is leveraged to compute the magnitude. This procedure allowed for precise and meaningful comparisons between various digit images within the dataset, without relying on segmentation results.

\setcounter{figure}{0}
\renewcommand{\figurename}{Figure B}
\begin{figure}[H]
  \includegraphics[width=4.80417in,height=2.43611in]{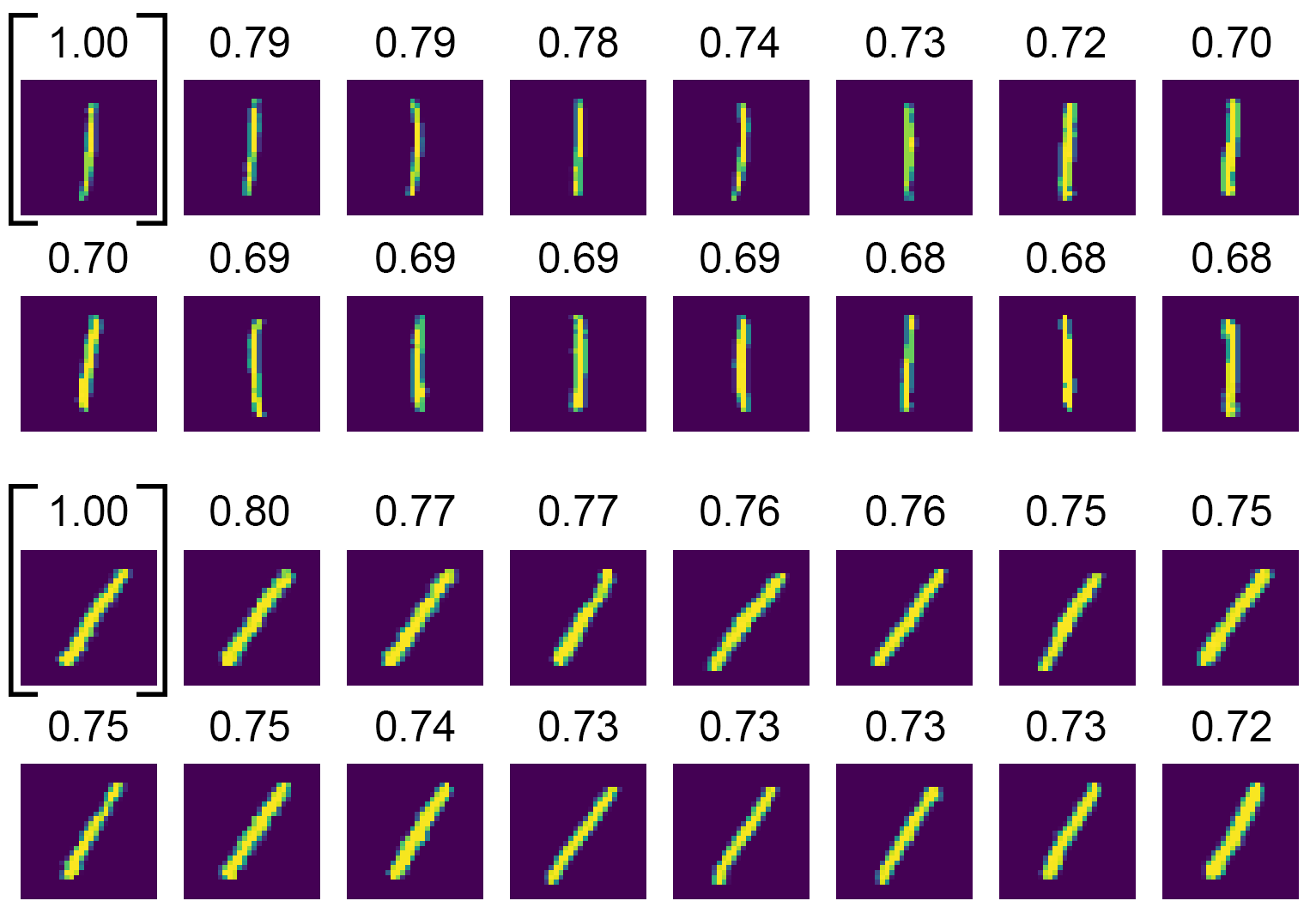}
  \caption{Examples of similarity results for the queried digit “1”, using a Gaussian membership function and ResNet50 as feature extractor. Average values are presented in decreasing order, from top left to bottom right. Note the difference of digit orientation between the results of the two similarity queries.}
\end{figure}

\begin{figure}[H]
  \includegraphics[width=3.31074in,height=2.29139in]{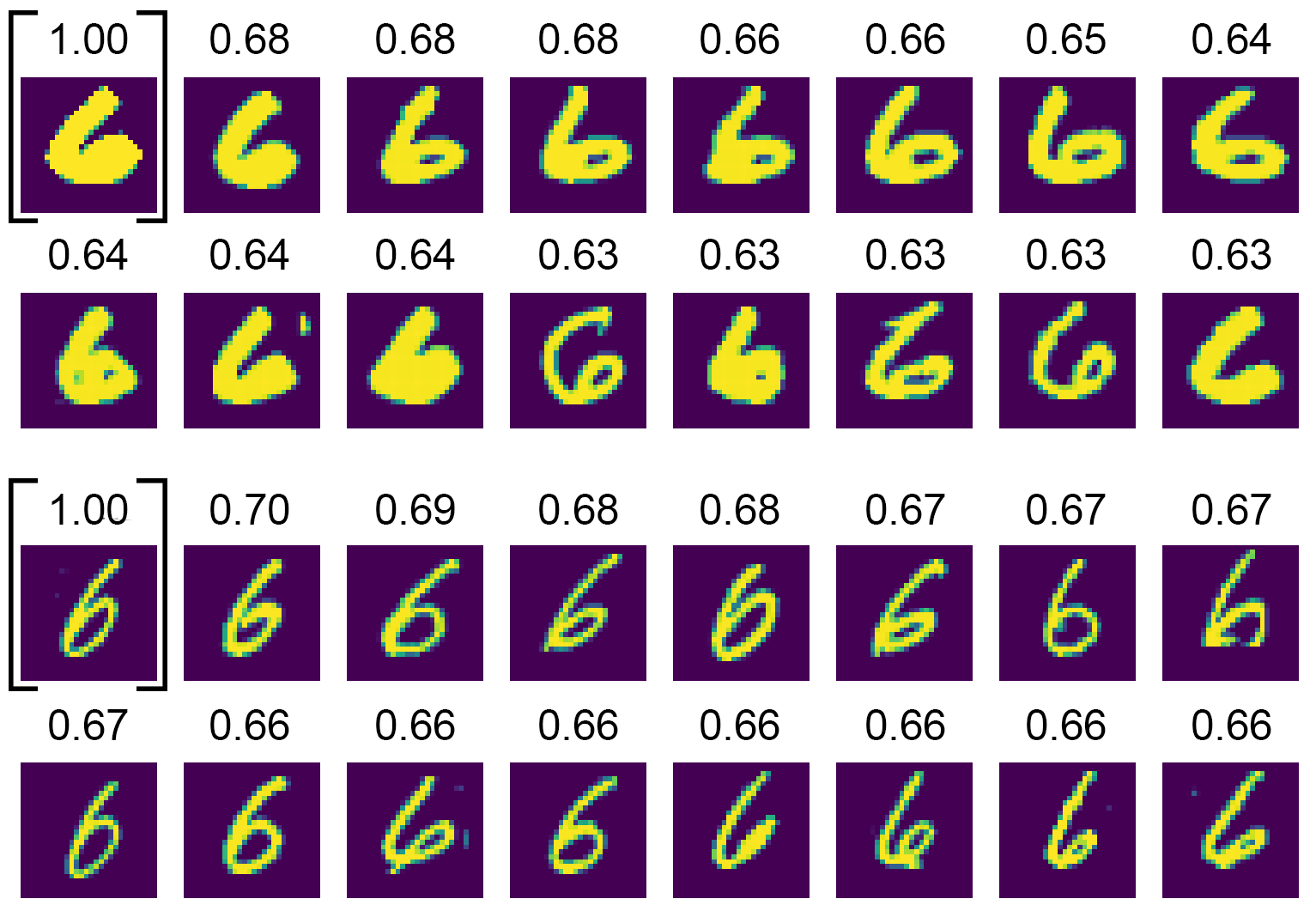}
  \caption{Examples of similarity results for the queried digit “6”, using a Gaussian membership function and ResNet50 as feature extractor. Average values are presented in decreasing order, from top left to bottom right. Note the difference of line thickness between the results of the two similarity queries.}
  \label{Fig.B2}
\end{figure}

\section{References}

\begin{enumerate}
\def\labelenumi{\arabic{enumi}.}
\item
  J. Wang, Y. Song, T. Leung, C. Rosenberg, J. Wang, J. Philbin, B. Chen
  and Y. Wu, ``Learning fine-grained image similarity with deep
  ranking'', in Proceedings of the IEEE Conference on Computer Vision
  and Pattern Recognition, pp. 1386-1393, 2014.
\item
  S. Hanif, C. Li, A. Alazzawe and L. J. Latecki, ``Image Retrieval with
  Similar Object Detection and Local Similarity to Detected Objects'',
  Pacific Rim Int. Conf. on Artificial Intelligence (PRICAI), Yanuca
  Island, Fiji, 2019.
\item
  Z. Ma, F. Liu, J. Dong, X. Qu, Y. He, S. Ji, ``Hierarchical Similarity
  Learning for Language-based Product Image Retrieval'', in ICASSP 2021
  - 2021 IEEE International Conference on Acoustics, Speech and Signal
  Processing (ICASSP), 2021.
\item
  S.-H. Cha, ``Comprehensive Survey on Distance/Similarity Measures
  between Probability Density Functions'', International Journal of
  Mathematical Models and Methods in Applied Sciences, 1 (4) pp.
  300-307, 2007.
\item
  C. Szegedy, W. Liu, Y. Jia, P. Sermanet, S. Reed, D. Anguelov, D.
  Erhan, V. Vanhoucke and A. Rabinovich, ``Going deeper with
  convolutions'', in IEEE Conference on Computer Vision and Pattern
  Recognition (CVPR), Boston, MA, 1-9, 2015.
\item
  A. Krizhevsky, I. Sutskever and G.E. Hinton, ``ImageNet Classification
  with Deep Convolutional Neural Networks'', Adv. Neural Inf. Process.
  Syst., 25, 2012.
\item
  G. Koch, R. Zemel, R. Salakhutdinov, ``Siamese Neural Networks for
  One-shot Image Recognition'', in ICML deep learning workshop, vol. 2,
  2015.
\item
  Y. Benajiba, J. Sun, Y. Zhang, L. Jiang, Z. Weng, O. Biran, ``Siamese
  Networks for Semantic Pattern Similarity'', in IEEE 13th International
  Conference on Semantic Computing (ICSC), 2019.
\item
  L. Nanni, G. Minchio, S. Brahnam, G. Maguolo, A. Lumini, ``Experiments
  of Image Classification Using Dissimilarity Spaces Built with Siamese
  Networks'', Sensor, 21(5), 1573, 2021.
\item
  A. Liao, M. Y. Yang, N. Zhan, B. Rosenhahn, ``Triplet-based Deep
  Similarity Learning for Person Re-Identification'' in IEEE
  International Conference on Computer Vision Workshops (ICCVW), 2017.
\item
  X. Yuan, Q. Liu, J. Long, L. Hu and Y. Wang, ``Deep Image Similarity
  Measurement Based on the Improved Triplet Network with Spatial Pyramid
  Pooling'', Information, 10(4), 129, 2019.
\item
  K. Shall, K. U. Barthel, N. Hezel and K. Jung, ``Deep Metric Learning
  using Similarities from Nonlinear Rank Approximation'', in IEEE 21st
  International Workshop on Multimedia Signal Processing (MMSP), 2019.
\item
  H. O. Song, Y. Xiang, S. Jegelka, S. Savarese, ``Deep Metric Learning
  via Lifted Structured Feature Embedding'', in IEEE Conference on
  Computer Vision and Pattern Recognition (CVPR), 2016.
\item
  J. Long, E. Shelhamer, and T. Darrell, ``Fully convolutional networks
  for semantic segmentation'', IEEE Conference on Computer Vision and
  Pattern Recognition (CVPR), 2015.
\item
  C. Juliani and E. Juliani, ``Deep Learning of Terrain Morphology and
  Pattern Discovery via Network-based Representational Similarity
  Analysis for Deep-Sea Mineral Exploration'', Ore Geology Reviews, 129,
  2020.
\item
  O. Ronneberger, P. Fischer, T. Brox, ``U-Net: convolutional networks
  for biomedical image segmentation'', MICCAI 2015: Medical Image
  Computing and Computer-Assisted Intervention -- MICCAI, pp: 234-241,
  2015.
\item
  A. Yu and K. Grauman, "Fine-Grained Visual Comparisons with Local
  Learning", in CVPR \textquotesingle14: Proceedings of the 2014 IEEE
  Conference on Computer Vision and Pattern Recognition, pp. 192--199,
  2014.
\item
  K. He, X. Zhang, S. Ren, J. Sun, ``Deep Residual Learning for Image
  Recognition'', in IEEE Conference on Computer Vision and Pattern
  Recognition, 2016.
\item
  D.P. Kingma and J.L. Ba, ``Adam: A method for stochastic
  optimization'', arXiv:1412.6980v9, 2015.
\item
  T. Gale, E. Elsen and S. Hooker, ``The state of sparsity in deep
  neural networks'', arXiv:1902.09574v1, 2019.
\item
  B. Liu, M. Wang, H. Foroosh, M. Tappen and M. Penksy, ``Sparse
  Convolutional Neural Networks'', in IEEE Conference on Computer Vision
  and Pattern Recognition (CVPR), 2015.
\item
  X. Xie, D. Du, Q. Li, Y. Liang, W.T. Tang, Z.L. Ong, M. Lu, H.P. Huynh
  and R.S.M. Goh, ``Exploiting Sparsity to Accelerate Fully Connected
  Layers of CNN-Based Applications on Mobile SoCs'', ACM Trans. Embed.
  Comput. Syst.17, 2, Article 37, 2017.
\item
  P.A. Golnari and S. Malik, ``Sparse matrix to matrix multiplication: A
  representation and architecture for acceleration'', arXiv:1906.00327v,
  2019.
\item
  Z. Wang, F. Li, G. Shi, X. Xie and F. Wang, ``Network pruning using
  sparse learning and genetic algorithm'', Neurocomputing, 404, 247-256,
  2020.
\item
  Y. Li, S. Lin, B. Zhang, J. Liu, D. Doermann, Y. Wu, F. Huang and R.
  Ji, ``Exploiting kernel sparsity and entropy for interpretable CNN
  compression'', arXiv:1812.04368, 2019.
\item
  S. Han, X. Liu, H. Mao, J. Pu, A. Pedram, M.A. Horowitz and W.J.
  Dally, ``EIE: Efficient Inference Engine on Compressed Deep Neural
  Network'', arXiv:1602.01528'', 2016.
\item
  S. Han, H. Mao and W.J. Dally, ``Deep compression: Compressing deep
  neural network with pruning, trained quantization and huffman
  coding'', arXiv:1510.00149, 2015.
\item
  W. Chen, J.T. Wilson, S. Tyree, K.Q. Weinberger and Y. Chen,
  ``Compressing Neural Networks with the Hashing Trick'',
  arXiv:1504.04788, 2015.
\item
  M.H. Zhu and S. Gupta, ``To prune, or not to prune: Exploring the
  efficacy of pruning for model compression'', arXiv:1710.01878v2, 2017.
\item
  B. Paria, C-K. Yeh, I.E.H. Yen, N. Xu, P. Ravikummar and B. Póczos,
  ``Minimizing FLOPs to learn efficient sparse representations'',
  arXiv:2004.05665v1, 2020.
\end{enumerate}

\end{document}